\documentclass[]{spie}  

\usepackage{subcaption}
\usepackage{amsmath,amsfonts,amssymb}
\usepackage{graphicx}
\usepackage[colorlinks=true, allcolors=blue]{hyperref}
\usepackage{booktabs}   
\usepackage{multirow}   
\usepackage{adjustbox}
\usepackage{authblk}

\title{DINOv3 with Test-Time Training for Medical Image Registration}

\author[,a]{Shansong Wang\textsuperscript{\#}}
\author[,a]{Mojtaba Safari\textsuperscript{\#}}
\author[a]{Mingzhe Hu}
\author[a]{Qiang Li}
\author[a]{Chih-Wei Chang}
\author[a]{Richard LJ Qiu} 
\author[a,*]{Xiaofeng Yang}

\affil[a]{Department of Radiation Oncology and Winship Cancer Institute, School of Medicine, Emory University, Atlanta, GA 30322}
\affil[ ]{\textsuperscript{\#} These authors contributed equally to this work.}

\begin{document}
\maketitle

\pagestyle{empty} 
\setcounter{page}{301} 

\begin{abstract}
Prior medical image registration approaches, particularly learning based methods, often require large amounts of training data, which constrains clinical adoption. To overcome this limitation, we propose a training-free pipeline that relies on a frozen DINOv3 encoder and test-time optimization of the deformation field in feature space. Across two representative benchmarks, the method is accurate and yields regular deformations. On Abdomen MR-CT, it attained the best mean Dice score (DSC) of 0.790 together with the lowest 95th percentile Hausdorff Distance (HD95) of \(4.9\pm5.0\) and the lowest standard deviation of Log-Jacobian (SDLogJ) of \(0.08\pm0.02\). On ACDC cardiac MRI, it improves mean DSC to \(0.769\) and reduces SDLogJ to \(0.11\) and HD95 to \(4.8\), a marked gain over the initial alignment. The results indicate that operating in a compact foundation feature space at test-time offers a practical and general solution for clinical registration without additional training.
\end{abstract}

\section{INTRODUCTION}
\label{sec:intro}  
Medical image registration is a key step in clinical diagnosis and scientific analysis~\cite{zitova2003image}. Aligning images from different time points, modalities, or patients enables the tracking of lesion progression, the integration of complementary multimodal information, and group-level statistical analysis~\cite{chen2025survey}. Traditional optimization-based~\cite{qiu2022embedding, pluim2003mutual} deformation models often rely on handcrafted similarity measures and optimization strategies. These approaches are computationally expensive and sensitive to noise and modality differences. With the growth of data scale and resolution, their limitations have become more pronounced, which has motivated the development of deep learning based approaches such as VoxelMorph~\cite{balakrishnan2019voxelmorph} and TransMorph~\cite{chen2022transmorph}. These models can directly predict deformation fields in an end-to-end manner, significantly improving speed and accuracy. However, they generally lack interpretability and still require manual segmentation as supervision in multimodal tasks to compensate for modality discrepancies.

To address these challenges, recent research has proposed extracting high-level semantic features with deep neural networks and optimizing correspondences in the feature space to enhance the robustness of registration~\cite{song2024dino}. Such methods benefit from the strong representational capacity of deep feature extractors, which capture rich contextual information and can be partially interpreted using techniques such as heatmaps~\cite{zhang2024utsrmorph,pan2025foundationmorph}. Nevertheless, directly applying generic deep features to medical image registration is not straightforward. The distributional gap between imaging modalities often reduces feature generalizability and necessitates modality-specific training~\cite{bian2025artificial}. Representative methods such as ConvexAdam~\cite{siebert2021fast} have demonstrated promising results in specific scenarios but typically rely on large-scale annotated data for optimization. Given the scarcity of annotated medical data in real clinical environments, the widespread adoption of these methods remains challenging.

Recently, self-supervised learning based vision foundation models have achieved breakthrough progress in natural images~\cite{oquab2023dinov2,pai2025vision,wang2025triad}. In particular, DINOv3 has shown that it can learn high-resolution image representations with remarkable quality without the need for manual labels, demonstrating strong potential for cross-task transfer~\cite{simeoni2025dinov3}. The combination of DINOv2~\cite{oquab2023dinov2} with test-time training (T\textsuperscript{3})~\cite{sun2020test} has been investigated in recent studies~\cite{gu2025vision,song2024dino}. This approach does not require fine-tuning of the feature extractor and instead optimizes only the deformation field during testing, achieving excellent performance in medical image registration. Such a strategy not only alleviates the problem of data scarcity but also meets the clinical demand for efficiency and reliability, offering a highly promising new pathway for medical image registration. However, the registration effect of the DINOv3~\cite{simeoni2025dinov3}+T\textsuperscript{3} scheme has not been explored yet.

In this study, we integrate DINOv3~\cite{simeoni2025dinov3} with test-time training to perform medical image registration without requiring any fine-tuning during training. Extensive experiments across multiple real-world medical image registration benchmarks demonstrate that our approach achieves superior performance and offers noteworthy clinical value for practical applications.

\section{METHOD}

\begin{figure}[!t] 
	\centering
	\includegraphics[width=\textwidth]{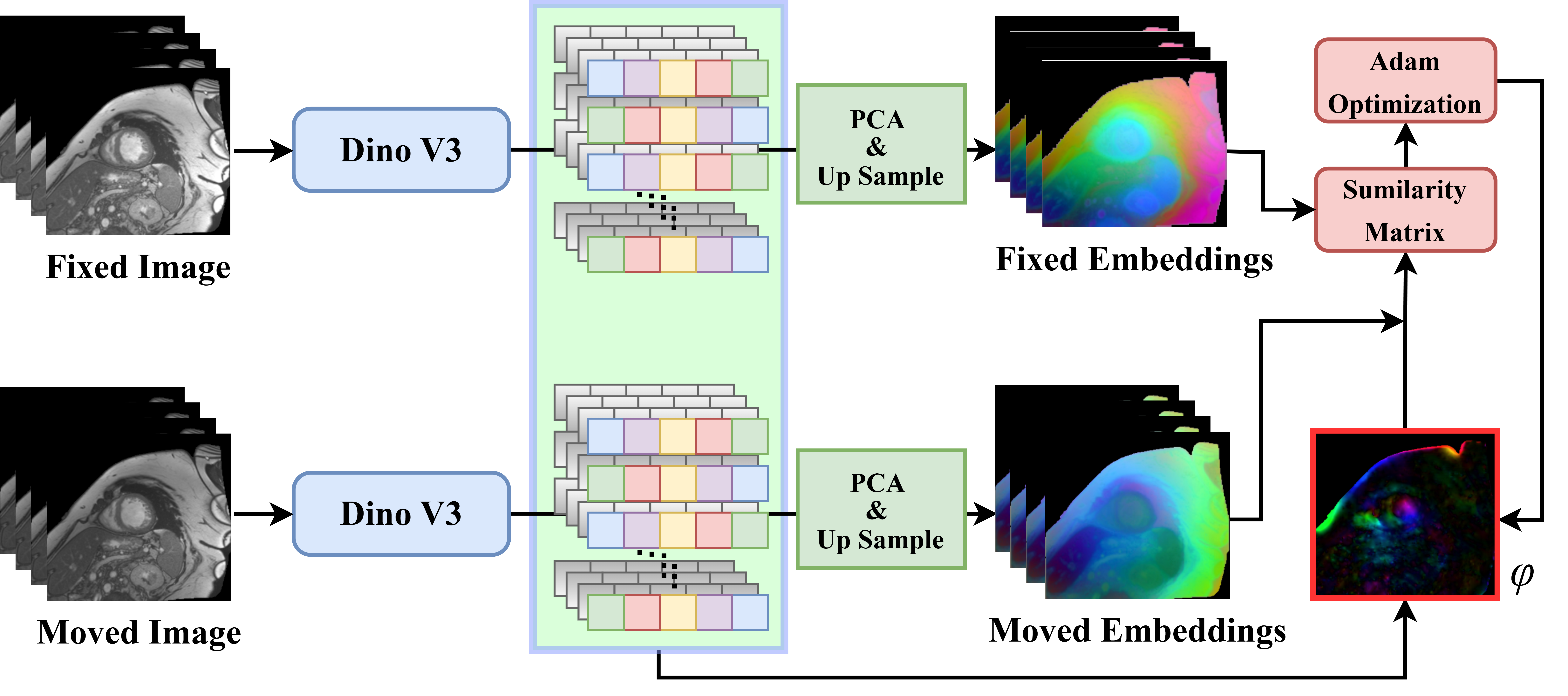}
	\caption{Overview of the proposed DINOv3+T\textsuperscript{3} pipeline for image registration.}
	\label{fig:f1} 
\end{figure}

We propose a training-free volumetric medical image registration pipeline that leverages a frozen DINOv3 vision encoder. The pipeline consists of three main stages: (1) slice-wise feature extraction using DINOv3, (2) dimensionality reduction through a joint feature bank with a shared projection, and (3) registration performed directly in the reduced feature space. Importantly, no fine-tuning of the encoder or additional supervision is required at any stage. An overview of the framework is provided in Figure~\ref{fig:f1}.

Let $I_{\text{fix}}$ and $I_{\text{mov}}$ denote the fixed and moving volumes, respectively. 
Since DINOv3 is inherently designed for 2D image inputs, each 3D volume is decomposed into a sequence of axial slices $\{I_{\text{fix}}^{z}\}_{z=1}^{Z}$ and $\{I_{\text{mov}}^{z}\}_{z=1}^{Z}$.  To balance computational efficiency with registration accuracy, only every $k$-th slice is passed through the encoder, where $k$ is a user-defined stride; in the special case $k{=}1$, all slices are processed.  From the encoder, we retain only the patch tokens, which capture localized semantic information within each slice.  Formally, the embedded representations are obtained as
\begin{equation}
\centering
    F_{\text{fix}}^{z} = E\!\left(I_{\text{fix}}^{z}\right),\qquad
    F_{\text{mov}}^{z} = E\!\left(I_{\text{mov}}^{z}\right),
\end{equation}
where $F^{z}\in\mathbb{R}^{N\times D}$, with $N$ denoting the number of patch tokens and $D$ the channel dimension (e.g., $D=1024$ for DINOv3-L). 
For slices that are skipped during encoding, missing features are reconstructed by linear interpolation of the token grids along the $z$-axis, thereby yielding complete feature sequences for both volumes.  These slice-wise feature maps serve as the foundation for constructing a joint feature bank, which enables subsequent dimensionality reduction and facilitates volumetric registration directly in the compact feature space.


Operating directly on high-dimensional embeddings can amplify noise and incur substantial computational cost. To address this, we compress the features into a shared low-dimensional subspace derived from a \emph{joint feature bank}. Specifically, all encoded tokens from the fixed and moving volumes are concatenated along the token dimension to form:
\begin{equation}
    F_{\text{all}}\in\mathbb{R}^{(2S\cdot N)\times D},
\end{equation}
where $S$ denotes the number of encoded slices per volume, $N$ the number of tokens per slice, and $D$ the channel dimension. Principal component analysis (PCA) is then applied to $F_{\text{all}}$, yielding a projection matrix $W_{\text{PCA}}\in\mathbb{R}^{D\times d}$ (with $d$ set to 24).  This projection is shared across volumes and applied slice-wise as:
\begin{equation}
\widetilde{F}_{\text{fix}}^{z}=F_{\text{fix}}^{z}W_{\text{PCA}},\qquad
\widetilde{F}_{\text{mov}}^{z}=F_{\text{mov}}^{z}W_{\text{PCA}},
\end{equation}
resulting in compact representations $\widetilde{F}^{z}\in\mathbb{R}^{N\times d}$. Each slice-level token matrix is then reshaped according to the underlying patch grid, producing a $(w\times h\times d)$ feature slice.  Stacking across $z$ yields a four-dimensional feature volume of size $W\times H\times Z\times d$, which is subsequently resampled to match the native voxel spacing of the input images. This ensures that the feature field and the image field are spatially aligned, enabling accurate volumetric registration in the reduced feature space.

Registration is performed in this reduced feature space. Let $\varphi:\Omega\rightarrow\mathbb{R}^{3}$ be a dense displacement field. We estimate $\varphi$ by minimizing the following loss function:
\begin{equation}
\mathcal{L}(\varphi)= \,\mathrm{Sim}\!\big(\widetilde{\mathbf{F}}_{\text{fix}},\,\widetilde{\mathbf{F}}_{\text{mov}}\circ\varphi\big)
\;+\;\lambda\|\nabla \varphi\|_2^{2},
\end{equation}
where $\widetilde{\mathbf{F}}_{\text{fix}},\widetilde{\mathbf{F}}_{\text{mov}}\in\mathbb{R}^{W\times H\times Z\times d}$. In our implementation, $\mathrm{Sim}$ is the mean-squared error between feature vectors at corresponding voxels, and $\lambda=1$ controls the smoothness regularization. The optimization proceeds in two complementary phases. A coarse-to-fine convex discrete search first explores a set of displacement candidates on multi-resolution grids to obtain a robust, approximately global solution $\varphi_{0}$ that captures large motions. Starting from this initialization, a continuous refinement based on Adam iteratively minimizes the same objective in a differentiable manner, enabling sub-voxel alignment while preserving regularity.

The proposed framework incorporates several design choices that provide practical flexibility. The slice stride parameter $k$ regulates the balance between computational efficiency and registration fidelity: setting $k=1$ ensures full volumetric coverage and obviates the need for interpolation, whereas larger values reduce runtime at the cost of finer detail. The reduced embedding dimension $d$ is selected to be sufficiently compact to enable stable optimization while retaining the representational capacity necessary for accurate alignment; increasing $d$ can recover additional structural detail, albeit with increased computational overhead. Although mean-squared error is employed as the default similarity measure, the formulation is inherently similarity-agnostic and can seamlessly integrate alternatives such as cosine similarity or normalized cross-correlation without altering the overall pipeline.  By combining frozen DINOv3 features, a shared low-dimensional embedding space, and a lightweight two-stage optimization scheme, the method achieves accurate and efficient 3D registration across diverse datasets without the need for training or annotated supervision.

\section{DATA ACQUSITION}

We validated our proposed registration method on 4D cardiovascular MRI dataset (unimodal) and abdomen MR-CT dataset (multi-modal).

\paragraph{1) Cardiovascular MRI} We evaluated our method on the unimodal 4D ACDC cardiovascular MRI test dataset~\cite{bernard2018deep}, which consists of 50 volumetric image pairs.  All volumes were resampled in the slice direction to 96 slices while preserving the original in-plane resolution, using linear interpolation implemented in the \texttt{SimpleITK} package. The corresponding segmentation contours were resampled to the same dimensions using nearest-neighbor interpolation. The ACDC dataset provides manual annotations of the left ventricle (LV), right ventricle (RV), and myocardium (Myo), which were used to quantitatively assess the performance of our method in comparison with baseline approaches.

\paragraph{2) Abdomen MR-CT} 
We further evaluated our method on the multi-modal Learn2Reg abdomen MR-CT dataset~\cite{hering2022learn2regcomprehensivemultitaskmedical}. 
We selected and curated representative samples for evaluation similar to the recently published study~\cite{song2024dino}. This dataset provides manual annotations for four organs including liver, spleen, left kidney (KidneyL), and right kidney (KidneyR), which were used to assess the performance of our method in comparison with baseline approaches.

\section{Evaluation}

We report Dice Similarity Coefficient (DSC), 95th Hausdorff distance (HD95), and the standard deviation of the log-Jacobian determinant (SDlogJ). For fixed mask $S_{\text{fix}}$ and warped moving mask $\hat S_{\text{mov}}=S_{\text{mov}}\!\circ\!\phi$, 
$\mathrm{DSC}=\frac{2|S_{\text{fix}}\cap \hat S_{\text{mov}}|}{|S_{\text{fix}}|+|\hat S_{\text{mov}}|}$. 
HD95 is the 95th-percentile symmetric surface distance between $\partial S_{\text{fix}}$ and $\partial \hat S_{\text{mov}}$, reported in voxels/mm. 
For deformation regularity, with $T(x)=x+\varphi(x)$ and $J(x)=|\nabla T(x)|$, we compute $\sigma_{\log J}=\operatorname{Std}_{x\in\Omega}[\log J(x)]$; smaller values indicate smoother, more plausible fields.

\section{RESULT}

On the Abdomen MR-CT dataset, our training free DINOv3+T\textsuperscript{3} attains the best mean DSC (0.790), surpassing the strongest DINO-Reg (0.780; \(+0.010\)) and all learning based competitors, while also delivering the lowest boundary error and deformation irregularity, with HD95 \(4.9\pm5.0\) (vs.\ \(7.3\pm10.1\) for the next best) and SDLogJ \(0.08\pm0.02\) (roughly \(40\%\) to \(60\%\) below the \(0.14\) to \(0.19\) range of baselines). Per organ, our method achieves the top spleen DSC \(0.770\pm0.13\) and nearly best liver performance \(0.809\pm0.10\) \((0.816\pm0.11\ \text{for DINO-Reg-Eco})\), while kidney DSCs trail DINO-Reg \((\mathrm{L}: 0.787\ \text{vs.}\ 0.856;\ \mathrm{R}: 0.780\ \text{vs.}\ 0.812)\). Overall, optimizing directly in a compact DINOv3 feature space at test time yields strong cross-modality alignment with sharper boundaries and more regular warps, and the remaining kidney gap suggests a promising direction for organ-aware weighting or adaptive similarity without sacrificing our training-free design.

On the ACDC cardiac MRI dataset, DINOv3+T\textsuperscript{3} surpasses DINOv2+T\textsuperscript{3} in overall and per-structure performance: mean DSC increases from 0.755 to 0.769 (LV 0.740 vs. 0.733, Myo 0.769 vs. 0.757, RV 0.797 vs. 0.774). Meanwhile, SDLogJ decreases from 0.16 to 0.11, a relative drop of about 34\%, and HD95 decreases from 5.1 to 4.8, a relative drop of about 6.9\%. Compared with the initial alignment, mean DSC improves by 0.176 and HD95 falls from 8.0 to 4.8, a reduction of about 40.4\%, which indicates that optimizing directly in the feature space at test time yields sharper boundaries and more regular deformations. 

\begin{table}[!b]
\centering
\caption{Quantitative registration results on the Abdomen MR-CT dataset. Reported values are averaged across the validation set. Arrows ($\uparrow$/$\downarrow$) indicate whether higher or lower values correspond to better performance.}
\label{tab:reg_results1}
\adjustbox{width=\textwidth,center}{
\begin{tabular}{cccccccc}
\hline
\multirow{2}{*}{Method} & \multicolumn{5}{c}{DSC$\uparrow$} & \multirow{2}{*}{SDLogJ$\downarrow$} & \multirow{2}{*}{HD95$\downarrow$} \\
\cmidrule(lr){2-6}
 & Mean & Liver & Spleen & KidneyL & KidneyR &  &  \\
\hline
Initial            & 0.371 & $0.509\pm0.15$ & $0.373\pm0.22$ & $0.297\pm0.20$ & $0.306\pm0.17$ & -          & $18.8\pm10.9$ \\
Voxelmorph         & 0.570  & $0.670\pm0.10$ & $0.566\pm0.26$ & $0.532\pm0.22$ & $0.512\pm0.28$ & $0.14\pm0.02$ & $13.6\pm10.3$ \\
Attention-Reg      & 0.556 & $0.651\pm0.01$& $0.536\pm0.25$ & $0.501\pm0.22$ & $0.537\pm0.27$ & $0.18\pm0.06$ & $14.2\pm9.6$  \\
TransMorph         & 0.570  & $0.672\pm0.01$& $0.559\pm0.26$ & $0.514\pm0.24$ & $0.536\pm0.27$ & $0.16\pm0.02$ & $13.1\pm9.5$  \\
ConvexAdam(MIND)   & 0.722 & $0.782\pm0.14$ & $0.649\pm0.31$ & $0.757\pm0.19$ & $0.700\pm0.29$ & $0.14\pm0.01$ & $8.9\pm10.9$ \\
DINO-Reg           & 0.780  & $0.777\pm0.10$ & $0.694\pm0.26$ & $\mathbf{0.856\pm0.04}$ & $\mathbf{0.812\pm0.12}$ & $0.17\pm0.02$ & $7.6\pm10.3$ \\
DINO-Reg-Eco       & 0.772 & $\mathbf{0.816\pm0.11}$ & $0.717\pm0.27$ & $0.836\pm0.06$ & $0.718\pm0.28$ & $0.19\pm0.04$ & $7.3\pm10.1$ \\
\hline
DINOv3+T\textsuperscript{3}       & $\mathbf{0.790}$ & $0.809\pm0.10$ & $\mathbf{0.770\pm0.13}$ & $0.787\pm0.09$ & $0.780\pm0.14$ & $\mathbf{0.08\pm0.02}$ & $\mathbf{4.9\pm5.0}$ \\
\hline
\end{tabular}}
\end{table}
\begin{table}[!b]
\centering
\caption{ACDC 4D cardiac MRI dataset results. Reported values are averaged across the validation set. Arrows ($\uparrow$/$\downarrow$) indicate whether higher or lower values correspond to better performance.}
\label{tab:reg_results2}
\adjustbox{width=\textwidth,center}{
\begin{tabular}{ccccccc}
\hline
\multirow{2}{*}{Method} & \multicolumn{4}{c}{DSC$\uparrow$} & \multirow{2}{*}{SDLogJ$\downarrow$} & \multirow{2}{*}{HD95$\downarrow$} \\
\cmidrule(lr){2-5}
 & Mean & LV & Myo & RV  &  &  \\
\hline
Initial             & $0.593\pm0.17$ & $0.662\pm0.11$ & $0.492\pm0.16$ & $0.625\pm0.19$   & - & $8.0\pm3.4$ \\
\hline
DINOv2+T\textsuperscript{3}      & $0.755 \pm 0.11$ & $0.733 \pm 0.11$ & $0.757 \pm 0.07$ & $0.774 \pm 0.15$ & $0.16 \pm 0.09$ & $5.1 \pm 3.7$ \\
DINOv3+T\textsuperscript{3}       & $\mathbf{0.769 \pm 0.10}$ & $\mathbf{0.740 \pm 0.11}$ & $\mathbf{0.769 \pm 0.08}$ & $\mathbf{0.797 \pm 0.12}$ & $\mathbf{0.11 \pm 0.06}$ & $\mathbf{4.8 \pm 3.5}$ \\
\hline
\end{tabular}}
\end{table}

\begin{figure}[!t] 
	\centering
	\includegraphics[width=\textwidth]{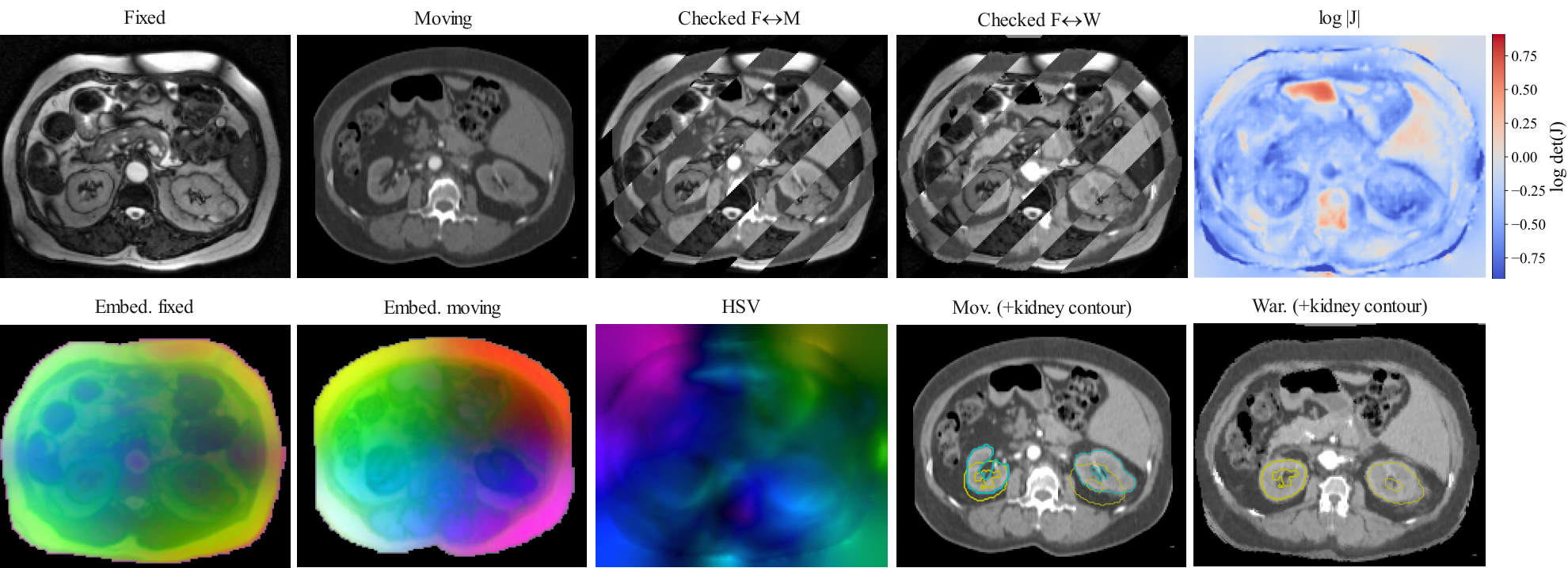}
    \caption{Qualitative results of the registration framework on MR–CT data. \textbf{Top row (left to right):} fixed image; moving image; checkerboard overlay (Fixed$\leftrightarrow$Moving); checkerboard overlay (Fixed$\leftrightarrow$Warped); and the $\log|J|$ map indicating deformation regularity. 
    \textbf{Bottom row (left to right):} DINOv3 embedding features of the fixed image; DINOv3 embedding features of the moving image; HSV composite of the moving image with kidney contour overlay; moving image with kidney contour overlay; and warped image with kidney contour overlay.}    \label{fig:registration_mrct}
\end{figure}

\begin{figure}[!t] 
	\centering
	\includegraphics[width=\textwidth]{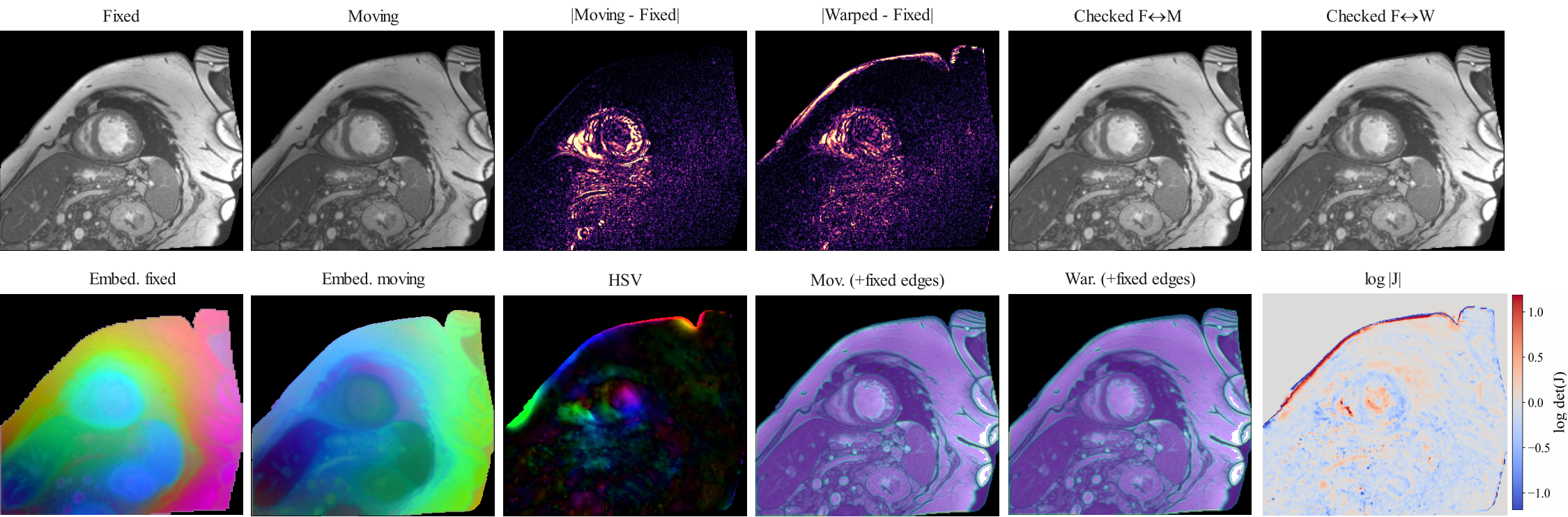}
    \caption{Qualitative results of the registration framework on ACDC cardiac MRI data.  \textbf{Top row (left to right):} fixed image; moving image; absolute difference $|\mathrm{Moving}-\mathrm{Fixed}|$; absolute difference $|\mathrm{Warped}-\mathrm{Fixed}|$; checkerboard overlay (Fixed$\leftrightarrow$Moving); checkerboard overlay (Fixed$\leftrightarrow$Warped). 
\textbf{Bottom row (left to right):} DINOv3 embedding features of the moving image; DINOv3 embedding features of the fixed image; HSV composite of the moving image with fixed-image edges; moving image with fixed-image edges; warped image with fixed-image edges; and the $\log|J|$ map illustrating deformation regularity.}

    \label{fig:registration_acdc}
\end{figure}

In addition to the quantitative improvements, qualitative results provide further insights into the effectiveness of the proposed method. Figure~\ref{fig:registration_mrct} presents cases from the abdominal MR-CT dataset. Organ boundaries, particularly around the kidneys, are more accurately aligned, and the overlay with contours demonstrates improved correspondence across modalities. The reduced boundary errors observed here are consistent with the superior HD95 and DSC scores in Table~\ref{tab:reg_results1}. Figure~\ref{fig:registration_acdc} shows results on the ACDC cardiac MRI dataset, where ventricular structures exhibit sharper definition and reduced mismatch in the difference maps. The checkerboard overlays further highlight improved alignment of myocardial borders, and the deformation fields remain regular, in line with the substantial reductions in HD95 and SDLogJ values reported in Table~\ref{tab:reg_results2}. 
Together, these qualitative examples reinforce the quantitative evidence that DINOv3+T\textsuperscript{3} achieves robust alignment across different anatomical regions and modalities, producing sharper boundaries, lower residual errors, and smoother deformations than competing approaches.

\section{CONCLUSION}
We introduced a training-free registration framework that combines a frozen DINOv3 encoder with test-time optimization of the deformation field in feature space. Across two benchmarks, our training-free framework improves accuracy and regularity. On Abdomen MR CT it achieves the best mean DSC 0.790 with the lowest HD95 \(4.9\pm5.0\) and SDLogJ \(0.08\pm0.02\). On ACDC, DINOv3+T\textsuperscript{3} outperforms DINOv2+T\textsuperscript{3} with mean DSC \(0.769\) and lower SDLogJ \(0.11\) and HD95 \(4.8\), markedly better than the initial alignment. Taken together across cross-modality abdomen and cardiac datasets, the framework consistently increases overlap accuracy while lowering boundary error or deformation irregularity without encoder fine-tuning or supervision, which supports its practicality in heterogeneous clinical environments.

\section{NOVELTY/BREAKTHROUGH}
To our knowledge, this is the first training-free medical image registration framework that pairs a frozen DINOv3 encoder with test time deformation optimization in a shared low-dimensional feature subspace learned via joint PCA, delivering modality robust registration that improves overlap, lowers HD95, and reduces SDLogJ across Abdomen MR CT, and ACDC.

\section{SUBMISSION STATUS}
This work has not been published or presented elsewhere.

\acknowledgments 
This research is supported in part by the National Institutes of Health under Award Number R56EB033332, R01EB032680, R01DE033512 and R01CA272991.

\bibliography{report} 
\bibliographystyle{spiebib} 

\end{document}